\pdfoutput=1
\documentclass[10pt,twocolumn,letterpaper]{article}

\usepackage[pagenumbers]{cvpr} % To force page numbers, e.g. for an arXiv version

\usepackage{xcolor}      % for colors
\usepackage[table]{xcolor}

\usepackage{tcolorbox}   % for colored boxes
\usepackage{multirow}
\usepackage{makecell}
\usepackage[table]{xcolor}
\usepackage{booktabs}

\definecolor{cvprblue}{rgb}{0.21,0.49,0.74}
\usepackage[pagebackref,breaklinks,colorlinks,allcolors=cvprblue]{hyperref}

\def\paperID{11822} % *** Enter the Paper ID here
\def\confName{CVPR}
\def\confYear{2026}

\title{Seeing Clearly, Reasoning Confidently: Plug-and-Play Remedies\\ for Vision Language Model Blindness}

\author{Xin Hu$^{1}$, Haomiao Ni$^{2}$, Yunbei Zhang$^1$, Jihun Hamm$^1$, Zechen Li$^1$, Zhengming Ding$^{1}$\\
$^1$Department of Computer Science, Tulane University,\\
$^2$Department of Computer Science, University of Memphis\\
}

\begin{document}
\maketitle
\begin{abstract}
Vision language models (VLMs) have achieved remarkable success in broad visual understanding, yet they remain challenged by object-centric reasoning on rare objects due to the scarcity of such instances in pretraining data. While prior efforts alleviate this issue by retrieving additional data or introducing stronger vision encoders, these methods are still computationally intensive during finetuning VLMs and don't fully exploit the original training data. In this paper, we introduce an efficient plug-and-play module that substantially improves VLMs' reasoning over rare objects by refining visual tokens and enriching input text prompts, without VLMs finetuning. Specifically, we propose to learn multi-modal class embeddings for rare objects by leveraging prior knowledge from vision foundation models and synonym-augmented text descriptions, compensating for limited training examples. These embeddings refine the visual tokens in VLMs through a lightweight attention-based enhancement module that improves fine-grained object details. In addition, we use the learned embeddings as object-aware detectors to generate informative hints, which are injected into the text prompts to help guide the VLM’s attention toward relevant image regions. Experiments on two benchmarks show consistent and substantial gains for pretrained VLMs in rare object recognition and reasoning. Further analysis reveals how our method strengthens the VLM's ability to focus on and reason about rare objects.

\end{abstract}    
\section{Introduction}
\label{sec:intro}

\begin{figure}[t]
  \centering
\includegraphics[width=\linewidth]{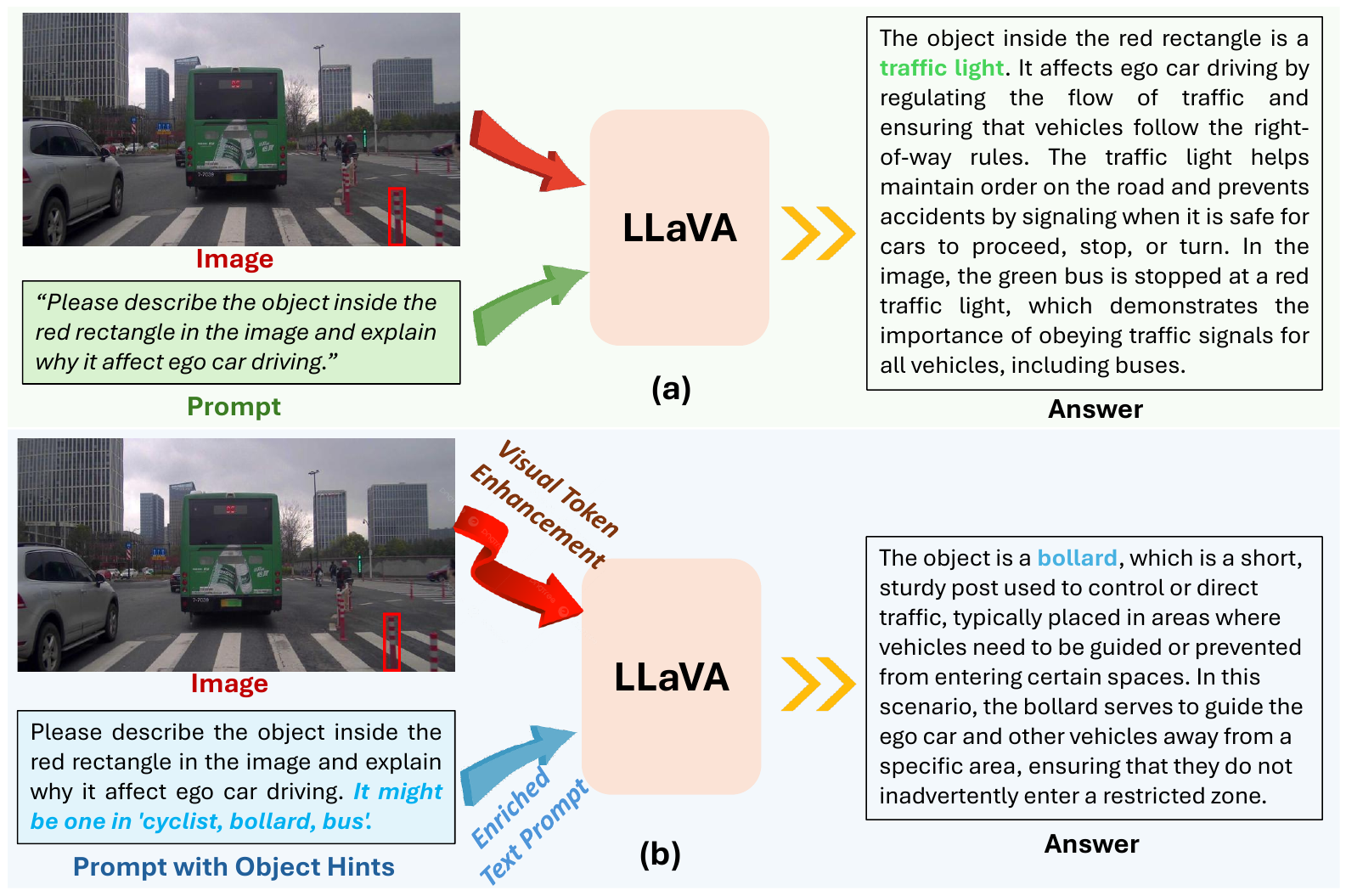}\vspace{-3mm}
  \caption{Comparison on rare object recognition: (a) shows that LLaVA tends to predict the ``bollard'' as a common object ``traffic light'',  while (b) demonstrates that our method corrects LLaVA by predicting ``bollard'' and providing reasoning through visual enhancement and text prompt enrichment with object hints, both based on the learned multi-modal class embeddings.
  % to correct recognition and reasoning without finetuning LLaVA. \JH{Quotation marks only in the above.} 
  % \JH{Would be nice if there is some self-explanatory illustration of what visual token enhancement is, or how class embedding is used.}
  }\vspace{-6mm}
  \label{fig:motivation}
\end{figure}

Vision language models (VLMs) have made remarkable advances in recent years, with both open-source \cite{bai2025qwen2, zhu2025internvl3, liu2023visual} and closed-source \cite{achiam2023gpt, team2023gemini} systems demonstrating strong performance across a wide range of multi-modal tasks. A key driver of this progress has been visual instruction tuning \cite{liu2023visual}, which bridges a pretrained vision encoder (e.g., CLIP \cite{radford2021learning}) and large language models via a lightweight projection layer. This design enables the language model to interpret and reason over visual inputs, thereby enabling effective vision-language alignment and fusion. Despite these successes, numerous studies \cite{tong2024cambrian, qi2025beyond, fu2025hidden} report persistent limitations of VLMs in vision-centric tasks such as referred object recognition and spatial reasoning. Particularly, VLMs perform much worse when dealing with rare or uncommon objects than common objects \cite{madan2024revisiting, robicheaux2025roboflow100, ahmed2025scaling}. For example, Figure~\ref{fig:motivation}(a) shows that LLaVA fails to recognize or reason correctly about the ``bollard,'' even when it is clearly visible in the input image. In contrast, our refinement on LLaVA resolves this issue, as illustrated in Figure~\ref{fig:motivation}(b).
% \JH{I like the idea of alignment and the overall approaches in this paper. Can we provide direct quantitative evidence that the failure to detect rare objects is due to the worse alignment with rare objects (compared to common objects)?}

% \XH{Existing problem from recent work}Despite these successes, VLMs frequently fail on object-centric visual reasoning on rare or uncommon objects \cite{}. As illustrated in Figure~\ref{fig:motivation}(a), VLMs are not able to recognize these objects or perform correct reasoning, even when they are clearly visible in the input image. This problem is critical in real-world applications such as autonomous driving, medical imaging, and robotics, where rare objects also carry significant semantic importance. \XH{Such problem is usually caused by the limited training data during VLM pretraining stage.} However, recent studies~\cite{viral,crossmodal} also criticize this weakness of visual representation in VLMs and attribute it to the indirect, language-mediated supervision used during pretraining, which shifts fine-grained, object-centric visual features from the original vision encoder to vague visual tokens in the VLM decoder.

Existing approaches largely attribute these shortcomings to the visual encoder or the projector. In response, subsequent works have introduced stronger vision encoders \cite{lu2024deepseek, li2023evaluating} and more expressive projectors \cite{liu2024improved, mckinzie2024mm1}, aiming to provide the language model with richer, more comprehensive visual representations. Recent studies \cite{gong2025kernel, yoon2025visual} leverage vision foundation models to align with the visual tokens in VLMs, making the visual tokens in VLMs preserve more spatial details during finetuning. While delivering measurable improvements, these methods are not specifically optimized toward rare objects, making them inefficient for such scenarios. \cite{liu2025few} attempts to mitigate the imbalanced distribution for rare objects through retrieval-augmented learning (RAL) from large-scale public data and builds a class-balanced training dataset. However, it still requires VLMs' computational finetuning and may lose original information. Based on these, it naturally raises the question:

\emph{How can we efficiently improve VLMs' capability in recognizing and reasoning about rare object-centric scenes?}

\begin{figure}[t]
  \centering
  \includegraphics[width=\linewidth]{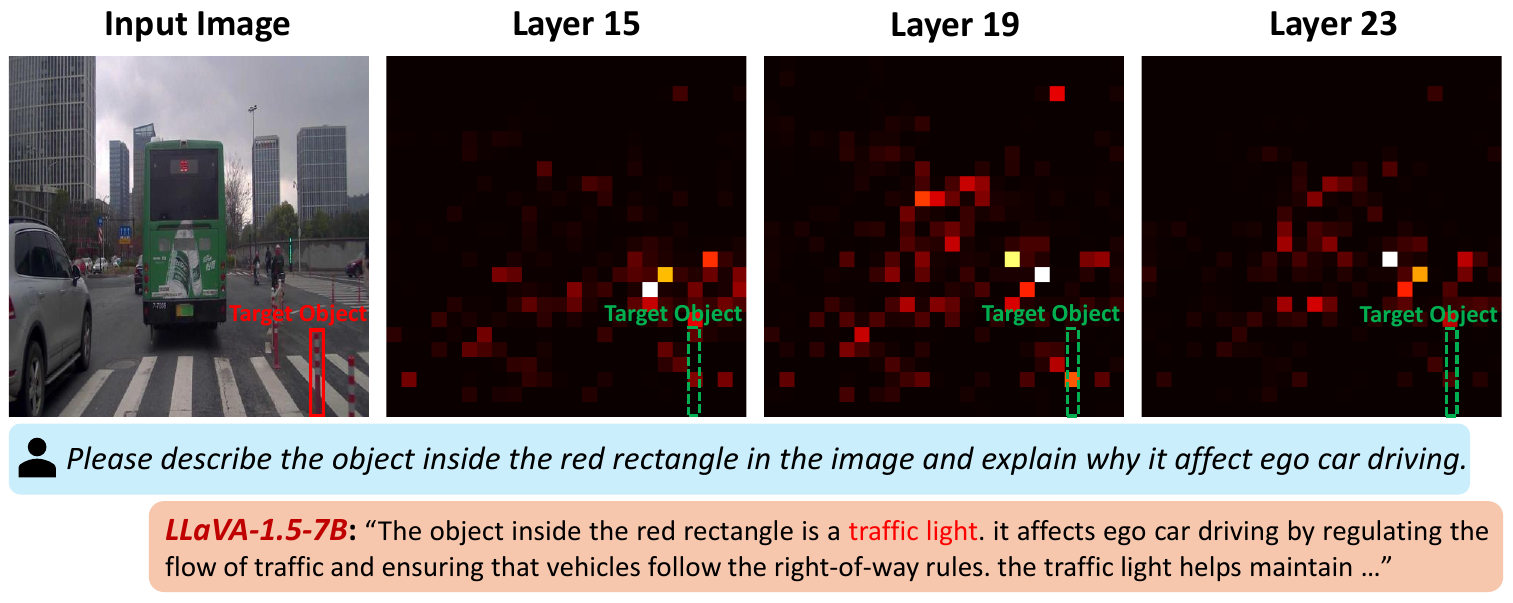}\vspace{-3mm}
  \caption{Visual attention on the object ``bollard'' from the CODA-LM dataset. The attention weights across layers show that LLaVA-1.5-7B allocates less attention to the target object region. Brighter colors indicate higher attention weights.}\vspace{-5mm}
  \label{fig:attention_analysis}
\end{figure}

To address this question, we first investigate \emph{``why VLMs struggle with rare, object-centric reasoning?''}. Recent studies \cite{zhang2025cross, jiang2025devils, yoon2025visual} suggest that VLMs primarily retrieve visual information in the middle decode layers when grounding referred visual objects. Inspired by this, we visualize attention weights on visual tokens in these layers of LLaVA-1.5-7B (Figure \ref{fig:attention_analysis}), highlighting how the predicted ``object'' tokens--``bollard'' attend to visual representations \cite{zhang2025cross}. We observe that LLaVA-1.5-7B focuses less on relevant object regions, leading to degraded reasoning performance. Based on these insights, we propose to mitigate such limitations from two complementary perspectives in the input level: \textit{(1) enhancing visual tokens in VLMs for rare objects, making them more salient for attention, and (2) guiding VLMs toward relevant object regions through enriched text prompts.}

In this paper, we propose an efficient plug-and-play module that empowers pretrained VLMs to see more clearly and reason more confidently for rare objects. Our core idea is to build multi-modal class embeddings for rare objects that merge the visual precision of foundation models with the semantic richness of synonym-augmented texts, creating powerful anchors for fine-grained, object-centric reasoning. Building on this, we introduce a \textbf{dual-mode} enhancement: first, we refine the visual tokens of VLMs via cross-attention with class embeddings, directly boosting object-level representations for more accurate and robust reasoning; second, we inject object hints in the text prompt by using class embeddings as detectors to infuse class-specific knowledge. To sum up, our main contributions are:
\begin{itemize}
\item We identify a critical blind spot of VLMs in reasoning over rare, object-centric scenes and propose an \emph{efficient module} built on learnable multi-modal class embeddings, enabling adaptation without finetuning VLMs.

% by  learnable multi-modal class embeddings that boost visual perception and enhance cross-modal reasoning with precision and efficiency. \JH{There is some finetuning in the proposed method, isn't there?}
% We identify and address the challenge of data scarcity and inadequate visual representations for rare objects in VLMs. A training-efficient solution via learnable class prototype embeddings is proposed to enable effective rare object reasoning without costly finetuning.
\item We introduce a \emph{dual-mode enhancement} framework that improves VLM reasoning from two complementary perspectives: \emph{visual token refinement} to sharpen object-level features, and \emph{text prompt enrichment with object hints} to enable more accurate object-centric reasoning.

% We introduce a \emph{dual-mode enhancement} framework that elevates VLM reasoning on two fronts: visual token refinement sharpens object-level features through cross-attention with class embeddings and object hints injection \XH{helps the model recognition and reasoning.}

% by guiding the attention weights.
% We introduce a dual-mode enhancement framework at semantic and representational levels. Text prior injection guides reasoning by constraining output space, while visual token refinement directly strengthens object-level features through cross-attention with class prototypes.
\item We conduct comprehensive evaluations on multiple challenging benchmarks, demonstrating significant performance gains, and further investigate how visual tokens and text hints enhance reasoning by interpreting the internal mechanisms of the language decoder.

% We evaluate on multiple challenging benchmarks for rare object reasoning with substantial gains, and also validate how visual token and text prior will improve the reasoning by interpreting the internal mechanism of language decoder.
% We provide comprehensive empirical validation across multiple challenging benchmarks for rare object recognition and reasoning. Extensive experiments demonstrate substantial improvements on different VLMs and provide insights into the mechanisms.
\end{itemize}

\section{Related Work}
\label{sec:related}

\noindent\textbf{Vision Language Models}: VLMs \cite{liu2023visual, bai2025qwen2, zhu2025internvl3} equip vision encoders \cite{radford2021learning} with large language models (LLMs) \cite{chiang2023vicuna, grattafiori2024llama} to form end-to-end systems that both perceive images and perform high-level reasoning. This design has driven strong performance on classic vision language tasks such as image captioning \cite{lin2014microsoft} and visual question answering (VQA) \cite{hudson2019gqa}. However, recent studies have shown that improvements in VQA tasks often stem from the strong language priors of the language model rather than from a precise perception of the input image \cite{fu2024blink, tong2024eyes, fu2025hidden, liu2025visual}. As a result, VLMs perform poorly on vision-centric benchmarks. \cite{madan2024revisiting, robicheaux2025roboflow100, ahmed2025scaling} particularly observe that VLMs' performance degrades sharply on rare object categories compared with common ones. 

% Inspired by these findings, we investigate how to efficiently improve VLMs' ability on rare-object-centric perception and reasoning.

\vspace{1mm}\noindent\textbf{Visual Improvement for VLMs}: To enhance the visual capability of VLMs, most prior efforts concentrate on the input stage--particularly the use of frozen vision encoders. These approaches have largely focused on adopting stronger vision encoders \cite{kar2024brave, shi2024eagle} or enhancing efficiency by reducing the overhead of visual tokens \cite{vasu2025fastvlm, yang2025visionzip}. Recent works \cite{yoon2025visual, gong2025kernel, zhang2025cross} have increasingly examined the internal information flow of VLMs and have taken a step further by advocating direct supervision of visual tokens in the middle layers, based on the finding that attention heads, concentrated in the middle layers, are pivotal for visual grounding. These advances have proven valuable, yet they don't mainly focus on object-level enhancement, especially for rare objects. Besides, these methods require sufficient data for VLM finetuning, which is not suitable for rare classes.

\vspace{1mm}\noindent\textbf{Training-free Adaptation of VLMs}: Training-free methods aim to adapt pretrained vision language models (VLMs) without finetuning, instead modifying inference while keeping all backbone weights frozen. Common approaches include score reweighting for zero-shot classification \cite{udandarao2023sus}, prototype-based similarity retrieval for task-specific predictions \cite{long2024training}, and compositional pipelines that leverage VLMs for complex reasoning and retrieval \cite{karthik2023vision}. Recent studies \cite{wu2024controlmllm, zhang2025towards} extend this idea by performing ``prompting in feature space''—injecting optimized latent prompts at test time to steer attention and decision boundaries without updating VLM parameters. Although yielding notable improvements under strict no-training constraints, these approaches struggle when VLMs are weak for rare objects. Our method follows this paradigm but differs by enhancing rare-object perception and reasoning by jointly refining visual token representations and textual prompts, keeping VLM frozen.
% Training-free methods seek to adapt pretrained VLMs without any finetuning, instead modifying inference while keeping all backbone weights frozen. Typical methods include score reweighting for zero-shot classification \cite{udandarao2023sus}, prototype-based similarity retrieval for task-specific predictions \cite{long2024training}, and compositional pipelines that utilize VLMs for complex reasoning and retrieval \cite{karthik2023vision}. Recent works \cite{wu2024controlmllm, zhang2025towards} perform ``prompting in feature space'' by injecting optimized latent prompts at test time, effectively steering attention and decision boundaries without updating VLM parameters. While these approaches report substantial gains under a strict constraint, they are not effective when the visual representations within VLMs are weak for rare object. Our work follows this paradigm but implements in a different way: enhancing rare-object-centric perception and reasoning in VLMs via multi-modal class embeddings via visual and text prompt refinement, while keeping the underlying VLM frozen.
\section{Proposed Method}

\subsection{Preliminaries}

VLMs integrate visual and textual information to enable multi-modal understanding. It typically includes three key components: a vision encoder $\mathcal{E}_\theta(\cdot)$, which extracts patch-level features $\mathbf{U} = \mathcal{E}_\theta(\mathbf{X})$ from an input image $\mathbf{X} \in \mathbb{R}^{H \times W \times 3}$ with height $H$ and width $W$; a connector $\mathcal{C}_\phi(\cdot)$, which maps visual features $\mathbf{U}$ into the language embedding space as $\mathbf{V} = \mathcal{C}_\phi(\mathbf{U})$; and a pretrained language model $\mathcal{L}_\psi(\cdot)$, which conducts auto-regressive next token generation based on the visual tokens $\mathbf{V} \in \mathbb{R}^{M \times D}$ and input text prompt which is first tokenized and embedded as $\mathbf{T} \in \mathbb{R}^{K \times D}.$ Note that $M$ is the number of visual tokens, $K$ is the length of text tokens, $D$ is the hidden dimension in $\mathcal{L}_\psi$($\cdot$); $\theta,\phi, \psi$ denote corresponding learnable parameters.

To generate answers, the visual and textual embeddings are firstly concatenated into a multi-modal sequence $\mathbf{S} = [\mathbf{V}; \mathbf{T}] \in \mathbb{R}^{(M+K) \times D}$ and pass through multiple transformer layers within $\mathcal{L}_\psi$($\cdot$). At each layer $\ell$, hidden states $\mathbf{H}^{(\ell)} = [\mathbf{H}_v^{(\ell)}; \mathbf{H}_t^{(\ell)}]$ are updated via a causal attention and feed-forward transformations, enabling interaction between visual and textual tokens. Generally, VLM is trained using a causal language modeling objective to finetune $\phi$ and $\psi$, learning to predict the next token conditioned on all preceding visual and textual tokens. This framework allows VLMs to perform tasks such as image captioning, visual question answering, and multi-modal dialogue. For our task, we mainly focus on the generated \textbf{``object'' token} for referred objects and relative reasoning answers.

\subsection{Motivation}

Previous works \cite{lu2024deepseek, mckinzie2024mm1, yoon2025visual, gong2025kernel} improve the visual token representation in VLMs by introducing stronger vision encoders or conducting alignment with the vision foundation model (VFM). These methods ignore object-level optimization, especially for rare objects. To address this, we propose class embeddings, which encode the essential characteristics of rare objects. Compared with others, these embeddings provide more class-specific information, thereby enhancing the model’s sensitivity to rare objects. Additionally, class embeddings can act as detectors, supplying object-aware knowledge that enriches input text prompts.

We will leverage the learnable multi-modal class embeddings as the information-enriched anchors via two complementary strategies: (1) visual token enhancement – enhancing visual token representations in VLM for richer multi-modal understanding; (2) text prompt enrichment – identifying potential objects using multi-modal class embeddings and incorporating them as hints into text prompts. By constructing and integrating multi-modal class embeddings, our approach enables VLMs to see clearly and reason confidently for rare object-centric understanding.

% enhance the object recognition by aligning visual tokens in VLMs with image features from vision foundation models (VFMs). This raises the question for our task: \emph{``How can a pretrained VLMs well handle for rare object recognition?''} In practice, rare objects often occupy small regions and contribute minimally to global representations, making such global alignment ineffective.

% \textbf{How can we provide more focused guidance to improve rare object recognition?} 

% \begin{figure}[t]
%   \centering
%   \includegraphics[width=\linewidth]{image/coda_lm_class_distribution.pdf}
%   \caption{\ZD{You can move to experiments as well} Class distribution and accuracy of CODA-LM dataset. It shows obvious imbalanced data across different classes, while our proposed method still improves the accuracy of most classes. \YZ{Only showing \#Instances is not informative, maybe add class-wise acc.} \JH{Our method is better for more common objects, the graph says.}}
%   \label{fig:distribution}
% \end{figure}

\begin{figure*}[t]
  \centering
\includegraphics[width=\linewidth]{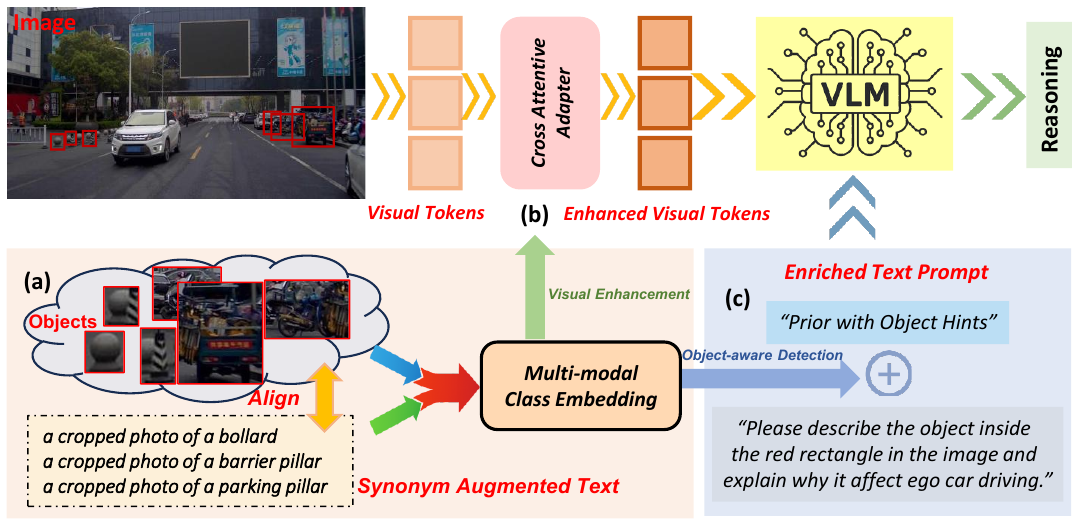}\vspace{-2mm}
  \caption{Overview of the model framework, which consists of three main components:
(a) a multi-modal class embedding learning module, which fuses object visual features with synonym-augmented text features;
(b) a visual token enhancement module, which applies a cross-attention mechanism between class embeddings and image visual tokens in VLMs; and
(c) a text hints injection module, which leverages the learned multi-modal class embeddings for object identification and enriches the text prompt with object hints.}\vspace{-4mm}
  \label{fig:framework}
\end{figure*}

\subsection{Learning Multi-modal Class Embedding}
\subsubsection{Adaptive Semantic Augmentation}

Building informative class embeddings is the key component in our methods. However, the limited, imbalanced data on rare objects make it hard to achieve this goal. 
% As shown in figure \ref{fig:distribution}, different rare object classes demonstrate extremely imbalanced data distribution, posing challenges for learning discriminative embeddings for all classes. 
To mitigate this issue, we first introduce textual augmentation to enrich the imbalanced data for achieving better class embeddings.

\vspace{1mm}\noindent\textbf{Semantic Enrichment.} For each rare object class $c \in \{1, \cdots, C\}$ with $N_c$ training samples, where $C$ is the total number of object class, we generate a set of text descriptions $\mathcal{T}_c$ using large language models (LLMs) such as ChatGPT \cite{achiam2023gpt}, Gemini \cite{team2023gemini}. The generation is performed in two complementary ways to improve the model's ability to learn robust class representations. Examples are shown as follows:
\begin{tcolorbox}[colback=pink!5!white, colframe=pink!80!black, title=\textbf{Lexical Variations}, rounded corners]
% Alternative names and synonyms capturing the same semantic concept with 

\textbf{Query:} \textcolor{green!60!black}{``\emph{Provide 5 alternative names or synonyms for the object class `stroller}'.''} \\
\textbf{Response:} \textcolor{orange!60!black}{``baby carriage''}, \textcolor{orange!60!black}{``pushchair''},...

% \textcolor{orange!60!black}{...}

% \textcolor{red!70!black}{``stroller''}, the generated variants include \textcolor{orange!60!black}{``baby carriage''} and \textcolor{orange!60!black}{``pushchair''}.  
\end{tcolorbox}

\begin{tcolorbox}[colback=orange!5!white, colframe=orange!20!white, title=\textbf{Visual Attributes}, rounded corners]
% Descriptive phrases highlighting visual characteristics with 

\textbf{Query:} \textcolor{green!60!black}{``\emph{List 20 phrases describing the visual attributes of a `stroller', including shape, material, and components.}''} \\
\textbf{Response:} \textcolor{orange!60!black}{``four-wheeled frame''}, \textcolor{orange!60!black}{``padded seat''},...
% For \textcolor{red!70!black}{``stroller''}, examples include \textcolor{orange!60!black}{``four-wheeled frame''} and \textcolor{orange!60!black}{``padded seat''}.  

\end{tcolorbox}

% This dual augmentation ensures that each class is represented both semantically and visually, improving the model's ability to learn robust class representations even with limited training samples.

% For the object class $c \in \{1, ..., C\}$, each class has different $N_c$ training samples, we generate a set of text descriptions $\mathcal{T}_c$ by utilizing LLMs like ChatGPT to augment the class descriptions from two ways: \ZD{Can you directly show two prompts?}

% \textit{Lexical Variations:} We collect alternative names and synonyms that capture the same semantic concept. For example, for the class ``stroller'', we include variations such as ``baby carriage'', ``pushchair''. This ensures the model learns that these different terms refer to the same visual concept.

% \textit{Visual Attributes:} We generate descriptive phrases highlighting visual characteristics including shape, color, material, and structural components. For ``stroller'', attributes might include ``four-wheeled frame'', ``padded seat'''. These descriptions help the model focus on the visual details.

\vspace{1mm}\noindent\textbf{Adaptive Augmentation.} To strengthen learning the class embeddings across imbalanced classes, we follow \cite{shi2023re} to re-sample generated textual descriptions for each class based on its visual sample (image) frequency. Specifically, the classes that have abundant visual samples receive a smaller range of textual variants, while rare classes are enriched with the most diverse descriptions.

\subsubsection{Visual-Language Alignment}

Although adaptive textual augmentation enhances semantic richness, it is also vital to capture fine-grained visual details. Therefore, we employ VFMs to extract visual representations for rare objects \cite{yoon2025visual, gong2025kernel} and jointly learn multi-modal class embeddings with visual and semantic supervision.

\vspace{1mm}\noindent\textbf{Dual Branch Feature Extraction.} As shown in Figure \ref{fig:framework} (a), we extract semantic features using a pretrained CLIP text encoder $\mathcal{F}_{\text{text}}$: $\mathbf{z}_t = \mathcal{F}_{\text{text}}(X_t) \in \mathbb{R}^{d_t}$, where $X_t \in \mathcal{T}_c$ and $d_t$ denotes the text feature dimension. For input visual image \textbf{X}, we first crop the object region image $\textbf{X}_{obj}$ with the bounding box and utilize a frozen VFM $\mathcal{F}_{\text{vis}}$($\cdot$) to capture object features $\mathbf{z}_v = \mathsf{AVG}(\mathcal{F}_{\text{vis}}(\textbf{X}_{obj})) \in \mathbb{R}^{d_v}$, where $d_v$ is the dimension and $\mathsf{AVG}(\cdot)$ is the global average pooling function. Both modalities are projected into language embedding space $\mathbb{R}^D$ of the language model $\mathcal{L}_\psi$($\cdot$):
{\small
\setlength{\abovedisplayskip}{4pt}%
  \setlength{\belowdisplayskip}{4pt}%
\begin{equation}
\mathbf{h}_t = \mathcal{G}_{\text{text}}(\mathbf{z}_t), \quad 
\mathbf{h}_v = \mathcal{G}_{\text{vis}}(\mathbf{z}_v),
\end{equation}}%
where $\mathcal{G}_{\text{text}}(\cdot)$ and $\mathcal{G}_{\text{vis}}(\cdot)$ are learnable MLP layers and $\mathbf{h}_v,\mathbf{h}_t \in \mathbb{R}^D$. To align multi-modal inputs, we first optimize $\mathcal{G}_{\text{text}}(\cdot)$ and $\mathcal{G}_{\text{vis}}(\cdot)$ with a cross-modal alignment loss to ensure consistency between modalities:
{\small
\setlength{\abovedisplayskip}{4pt}%
  \setlength{\belowdisplayskip}{4pt}%
\begin{equation}
\mathcal{L}_{\text{align}} = -\frac{1}{N}\sum_{i=1}^{N} \log 
\frac{\sum_{j \in \mathcal{P}_i} \exp(\langle \mathbf{h}_v^i, \mathbf{h}_t^j \rangle )}
{\sum_{o=1}^{|\mathcal{T}|} \exp(\langle \mathbf{h}_v^i, \mathbf{h}_t^o \rangle)},
\end{equation}}%
where $\langle \cdot, \cdot \rangle$ is the cosine similarity function, $\mathcal{P}_i$ indexes all augmented texts of the same class as $\mathbf{h}_v^i$, $|\mathcal{T}|$ is the total number of augmented texts for all classes, and $N$ is the total number of training samples. This encourages each projected visual feature to align with all its semantic variations.

% For visual images, a frozen vision foundation model $\mathcal{F}_{\text{vis}}$ is utilized to extract object features with Region of Interest (ROI) as $\mathbf{z}_v = \mathsf{AvgPool}(\mathsf{ROIAlign}(\mathcal{F}_{\text{vis}}(\mathbf{X}), \mathbf{B})) \in \mathbb{R}^{d_v}$, where $\mathbf{B}$ is object bounding box, $d_v$ is the visual feature dimension and $\mathsf{AvgPool}(\cdot)$ is the spatial average pooling function. This ensures the model focuses on the object region while leveraging rich visual representations.

% The resulting embeddings form class-specific prototypes that are both semantically informative and visually discriminative, supporting robust reasoning even with limited data.

\vspace{1mm}\noindent\textbf{Learning Multi-modal Class Embeddings.} After the marginal alignment, we establish a set of learnable class embeddings $\mathbf{W} = \{\mathbf{w}_1, \cdots, \mathbf{w}_C\}$, where $\mathbf{w}_c \in \mathbb{R}^D$ represents the embedding for class $c$. Instead of random initialization, the embedding begins from the averaged projected visual features as: $
\mathbf{w}_c^{(0)} = \mathsf{AVG}\Big(\mathbf{h}_v^{(c,i)}|_{i=1}^{N_c}\Big)$,
where $\mathbf{h}_v^{(c,i)}$ is the projected visual embedding of the $i$-th sample in class $c$. This will leverage the discriminative visual representations from VFM and also make the class embeddings more reliable at the beginning.

To effectively bridge the visual and textual modalities, we jointly optimize the textual generator $\mathcal{G}_{\text{text}}(\cdot)$, the visual generator $\mathcal{G}_{\text{vis}}(\cdot)$ according to Eqs.~(2) and (3) given the multi-modal class embeddings $\mathbf{W}$:
% contrastive learning loss to jointly and iteratively optimize the text and visual encoders, $\mathcal{G}_{\text{text}}$ and $\mathcal{G}_{\text{vis}}$, along with the class embeddings $\mathbf{W}$.
% \vspace{1mm}\noindent\textit{Class-Level Discrimination:} To ensure clear boundary for class embeddings, we align both visual and textual embeddings with class embedding $\mathbf{w}_c$ using a shared contrastive formulation:
{\small
\setlength{\abovedisplayskip}{4pt}%
  \setlength{\belowdisplayskip}{4pt}%
\begin{equation}
\mathcal{L}_{\text{class}} = -\frac{1}{N+|\mathcal{T}|} \sum_{\textbf{x}_c \in \{\mathbf{h}_v, \mathbf{h}_t\}} 
\log \frac{\exp(\langle \textbf{x}_c, \mathbf{w}_{c} \rangle)}{\sum_{j=1}^C \exp(\langle \textbf{x}_c, \mathbf{w}_j \rangle)},
\end{equation}}%
where $\textbf{x}_c$ denotes the projected embedding $\mathbf{h}_v,\mathbf{h}_t$ belonging to class $c$. This ensures both visual and textual features are discriminatively aligned with their class embedding. 

% \XH{
% simultaneously, we also incorporate the cross-modal alignment loss $\mathcal{L}_{\text{align}}$ to constrain the modal consistency.}

Iteratively, we update the multi-modal class embedding $\mathbf{w}_{c}$ with exponential moving average (EMA) policy:
% \begin{equation}
$\mathbf{w}_c^{(t+1)} = \kappa \cdot \mathbf{w}_c^{(t)} + (1-\kappa) \cdot \bar{\mathbf{h}}_v^{(c)},$
% \end{equation}
where $\bar{\mathbf{h}}_v^{(c)} = \mathbb{E}[\mathbf{h}_v^i\mid{y_i=c}]$ is the mean visual embedding for samples of class $c$ after optimization at time step $t$ and $\kappa$ is the momentum coefficient. This ensures stable updates to class embeddings to adapt to the alignment between visual and textual features.

\subsection{Visual Token Refined Perception}
% \ZD{please update this.}
% While some VLMs fundamentally lack sufficient visual representations for rare objects. This limitation not only stems from their pretrained datasets with similar long-tail distributions but also is caused by implicit text supervision. For these cases, we propose to utilize the class embedding to enhance the visual representations.

Existing works \cite{yoon2025visual, gong2025kernel} improve the visual tokens in VLMs by finetuning the whole vision language model together with a VFM. Although effective, this strategy is computationally expensive and risks catastrophic forgetting of the pretrained VLM. In contrast, we introduce a lightweight adapter that refines the frozen visual tokens $\mathbf{V}$ using the previously learned class embeddings $\mathbf{W}$. The adapter is trained to preserve useful information in the original $\mathbf{V}$ while injecting class-discriminative cues from $\mathbf{W}$.

% directly finetune the VLM with vision foundation models to improve the visual latent tokens, which is computationally expensive and risk catastrophic forgetting of pretrained knowledge. Instead, we introduce a lightweight adapter module that refines visual tokens while keeping the original VLM frozen. The adapter will preserve useful information from pretrained visual tokens while injecting discriminative features from the class embeddings.

% \noindent\textbf{Attentive Autoencoder.} As shown in Figure \ref{fig:framework} (b), we design an autoencoder with cross attention fusion to refine the original visual tokens $\mathbf{V}'$ in VLM with class embeddings $\mathbf{W}$. Specifically, the visual token will set as query, and class embedding are serving as key and value. To align the feature space, we introduce encoder and decoder to transfer the visual tokens $\mathbf{V}'$ into the feature space of class embeddings $\mathbf{W}$ and then decode to original dimension. Interestingly, if we directly conduct cross attention fusion between $\mathbf{V}'$ and $\mathbf{W}$, the performance is not as good as ours, as shown in Table \ref{}. Finally, the proposed module improves visual tokens $\mathbf{W}$ with a refined version $\hat{\mathbf{V}}'$.

\vspace{1mm}\noindent\textbf{Cross Attentive Adapter.} Given visual tokens $\mathbf{V} \in \mathbb{R}^{M \times D}$ from the frozen VLM and multi-modal class embeddings $\mathbf{W} \in \mathbb{R}^{C \times D}$, we introduce a cross attentive visual token adapter $\mathcal{A}_\omega(\cdot)$ with learnable parameters $\omega$ that outputs refined visual tokens $\hat{\mathbf{V}} \in \mathbb{R}^{M \times D}$:
{\small
\setlength{\abovedisplayskip}{4pt}%
  \setlength{\belowdisplayskip}{4pt}%
\begin{equation}
\hat{\mathbf{V}} 
= \mathcal{A}_\omega(\mathbf{V},\mathbf{W})
=  \mathbf{V} + \mathcal{C}_{\text{att}}\big(\mathbf{V}, \mathbf{W}
    \big),
\end{equation}}%
% where we design adapter $\mathcal{A}_\omega(\cdot)$ as a 
% $\mathcal{E}_{\text{enc}}(\cdot)$ is a \ZD{[what is this? It seems V' and W have the same dimension in Eq (8)] lightweight} encoder that maps $\mathbf{V}'$ into a latent feature space, 
where $\mathcal{C}_{\text{att}}(\cdot, \cdot)$ is a multi-head 
cross-attention module with the visual tokens $\mathbf{V}$ as queries and the class embeddings $\mathbf{W}$ as keys and values. This module explicitly enforces that the refinement of visual tokens is driven by the class-aware knowledge from multi-modal class embeddings. Since we keep VLMs frozen, it is important to ensure that the refined visual tokens $\hat{\mathbf{V}}$ have a distribution similar to that of the original visual tokens $\mathbf{V}$. To reach this, we encourage them to stay close:

% $\hat{\mathbf{V}}'$ to stay close to $\mathbf{V}'$:
% and $\mathcal{D}_{\text{dec}}(\cdot)$ decodes the fused features back to the original visual token space. 
{\small
\setlength{\abovedisplayskip}{4pt}%
  \setlength{\belowdisplayskip}{4pt}%
\begin{equation}
\mathcal{L}_{\text{rec}}
=\big\|
\mathcal{A}_\omega(\mathbf{V}, \mathbf{W}) - \mathbf{V}
\big\|_2^2.
\end{equation}}
% where $M$ is the number of visual tokens.\ZD{You define the size of V' before? N x D? and You used N in Eqs (2) \& (3). Please fix.}

% \noindent\textit{Autoregressive language modeling loss.} 
To ensure the refined visual tokens $\hat{\mathbf{V}}=\mathcal{A}_\omega(\mathbf{V}, \mathbf{W})$ are valid to generate correct reasons, we deploy the standard causal language modeling objective of the VLM as:
% Second, we train the adapter to support the downstream language generation using the refined visual tokens
% $\hat{\mathbf{V}}' = \mathcal{A}_\omega(\mathbf{V}', \mathbf{W})$ and text embeddings $\mathbf{T}'$:
{\small
\setlength{\abovedisplayskip}{4pt}%
  \setlength{\belowdisplayskip}{4pt}%
\begin{equation}
\mathcal{L}_{\text{autoreg}}
= - \sum_{i=1}^{K} 
\mathsf{log}_{p_\psi}\big(
\mathbf{T}_i \mid \mathbf{T}_{<i}, \mathcal{A}_\omega(\mathbf{V}, \mathbf{W})
\big),
\end{equation}}\\
where $\mathbf{T} \in \mathbb{R}^{K \times D}$ denotes the $K$ text tokens as stated before and $p_\psi(\cdot)$ is the output distribution of the frozen language model $\mathcal{L}_\psi$($\cdot$).

To this end, we optimize the adapter’s learnable parameters $\omega$ while keeping all other modules frozen, using the following joint objective:
{\small
\setlength{\abovedisplayskip}{4pt}%
  \setlength{\belowdisplayskip}{4pt}%
\begin{equation}
\mathcal{L}_{\text{adapter}}
= \mathcal{L}_{\text{rec}} + \mathcal{L}_{\text{autoreg}}.
\end{equation}}

\subsection{Text Hints Injected Reasoning}

% To strengthen the VLM's reasoning ability, we will augment the original test prompt with additional textual priors that steer the model's reasoning toward the relevant objects depicted in the images.
Beyond refining visual tokens with the attentive adapter, we further exploit the multi-modal class embeddings $\mathbf{W}$ to inject object-aware textual hints into the VLM's reasoning process. This realizes the second enhancement strategy: text prompt enrichment guided by class embeddings.

\vspace{1mm}\noindent\textbf{Object-Aware Detection.} We treat the learned class embeddings $\mathbf{W}$ as object-specific detectors. Given an input image $\mathbf{X}$, we reuse the frozen VFM $\mathcal{F}_{\text{vis}}$($\cdot$) and the visual projection head $\mathcal{G}_{\text{vis}}$($\cdot$) to obtain $M$ visual tokens. We then compute cosine similarities between these tokens and the class embeddings $\mathbf{W}$, producing a class-wise score map:
{\small
\setlength{\abovedisplayskip}{4pt}%
  \setlength{\belowdisplayskip}{4pt}%
\begin{equation}
\mathbf{S} 
= \mathsf{cos}\big(\mathcal{G}_{\text{vis}}(\mathcal{F}_{\text{vis}}(\mathbf{X})), \mathbf{W}\big) 
\in \mathbb{R}^{M \times C},
\end{equation}}where $\mathbf{S}_{i,c}$ denotes the similarity between the $i$-th visual token and the $c$-th class embedding. 
For each class $c$, we aggregate patch-level evidence into a global relevance score $r_c = \max_{1 \le i \le M} \mathbf{S}_{i,c}$ and select the top-$k$ classes with the highest scores as detected categories, which serve as the image-conditioned hints over candidate objects. 
% The resulting set of detected categories is denoted by $\mathcal{C}_{\text{top}}$ and serves as an image-conditioned prior over candidate objects.

% To identify potential object hints, we leverage the learned multi-modal class embeddings as object-specific detectors. Given an input image $\mathbf{X}$, we localize objects by computing the cosine similarity between $m$ extracted visual tokens and $C$ multi-modal class embeddings, resulting in a class-wise probability score map over all image patches:
% \begin{equation}
% \mathbf{S} = \mathsf{cos}\big(\mathcal{G}{\text{vis}}(\mathcal{F}{\text{vis}}(\mathbf{X})), \mathbf{W}\big) \in \mathbb{R}^{m\times{C}}.
% \end{equation}

% To emphasize potential object priors, we then select the top-$k$ classes based on the maximum row-sum scores in $\mathbf{S}$, and 

\begin{table*}[t]
\centering
\caption{
GPT score comparison on the CODA-LM dataset. 
``+ Ours'' denotes our parameter-efficient refinement built on frozen baseline VLMs. 
Models marked with $^\dagger$ are task-specific finetuned models on CODA-LM, and models marked with $^\ddagger$ are training-free methods. 
}\vspace{-2mm}
\begin{tabular}{c|cccccccc}
\Xhline{1pt}
\textbf{Model / Metrics}
& \textbf{Barrier}$\uparrow$ 
& \textbf{Other}$\uparrow$ 
& \textbf{Cone}$\uparrow$ 
& \textbf{Light}$\uparrow$ 
& \textbf{Sign}$\uparrow$ 
& \textbf{Vehicle}$\uparrow$ 
& \textbf{VRU}$\uparrow$ 
& \textbf{All}$\uparrow$ \\
\Xhline{1pt}

LLaVA-1.5-7B \cite{liu2023visual}             & 39.3 & 40.2 & 54.5 & 54.4 & 48.8 & 48.9 & 40.5 & 46.5 \\
\rowcolor{blue!8}
LLaVA-1.5-7B + Ours                           & 68.3 & 68.3 & 84.9 & 61.4 & 48.2 & 73.0 & 56.1 & 72.8 \\
Qwen2.5-VL-7B \cite{bai2025qwen2}            & 70.9 & 62.5 & 84.9 & 48.8 & 52.1 & 66.5 & 54.6 & 67.9 \\
\rowcolor{blue!8}
Qwen2.5-VL-7B + Ours                         & \textbf{79.8} & \textbf{73.8} & \textbf{91.7} & 64.3 & 54.3 & 71.0 & 58.4 & 75.4 \\
InternVL3-8B \cite{zhu2025internvl3}         & 59.7 & 65.5 & 73.3 & 64.4 & 52.3 & 66.9 & 59.6 & 65.4 \\
\rowcolor{blue!8}
InternVL3-8B + Ours                          & 76.4 & \underline{69.3} & 85.8 & \underline{67.5} & \underline{55.1} & 73.8 & 66.2 & 74.2 \\
LLaVA-1.5-13B \cite{liu2023visual}           & 40.9 & 41.8 & 46.3 & 58.8 & 47.7 & 61.0 & 41.5 & 50.8 \\
\rowcolor{blue!8}
LLaVA-1.5-13B + Ours                         & 70.1 & 59.6 & \underline{87.1} & 53.1 & 51.4 & 73.4 & 57.5 & 71.2 \\ \hline

\rowcolor{gray!10}
CODA-LM$^\dagger$ \cite{chen2025automated}   & \underline{78.7} & 68.8 & 86.2 & \textbf{73.3} & \textbf{64.9} & \underline{78.8} & \textbf{73.8} & \textbf{77.7} \\
\rowcolor{gray!10}
MiniDrive$^\dagger$ \cite{zhang2024minidrive} & 62.9 & 62.8 & 84.4 &  --  &  --  & 67.4 & 36.0 & 66.3 \\
\rowcolor{gray!10}
MPDrive$^\dagger$ \cite{zhang2025mpdrive}    & 70.0 & 62.8 & 77.7 &  --  &  --  & \textbf{79.5} & \underline{70.0} & \underline{76.1} \\ \hline

\rowcolor{green!5}
Jiang et al.$^\ddagger$ \cite{jiang2025devils}  
                          & 40.3 & 41.4 & 52.1 & 60.8 & 45.9 & 49.4 & 43.1 & 46.7 \\
\rowcolor{green!5}
ControlMLLM++$^\ddagger$~\cite{wu2024controlmllm} 
                          & 39.3 & 45.4 & 53.0 & 62.2 & 46.6 & 50.1 & 39.0 & 47.0 \\
\Xhline{1pt}
\end{tabular}\vspace{-3mm}
\label{tab:coda-lm}
\end{table*}

We then augment the original input prompt with these candidate objects as follows:
\begin{tcolorbox}[colback=orange!5!white, colframe=orange!20!white, title=\textbf{Text Hints Injected Prompt}, rounded corners]
\textbf{Original prompt:} \textcolor{red!60!black}{``\emph{Please describe the object inside the red rectangle in the image.}''}

\textbf{New prompt:} \textcolor{blue!50!black}{``\emph{Please describe the object inside the red rectangle in the image. [Detected: \{top-$k$ classes\}]}}''
\end{tcolorbox}

At inference time, we feed both the refined visual tokens $\hat{\mathbf{V}} 
= \mathcal{A}_\omega(\mathbf{V}, \mathbf{W})$ from the attentive adapter, and the embeddings of the text hints injected prompt into the frozen language model $\mathcal{L}_\psi$($\cdot$). This produces a complementary synergy: the adapter $\mathcal{A}_\omega(\cdot)$ provides richer visual features that capture fine-grained characteristics of rare objects, while the text hints explicitly guide the language model to focus on these objects and interpret the enhanced representations more accurately. Importantly, this mechanism is computationally efficient: it only requires updating the lightweight adapter and class embeddings, adapts to different VLM backbones, and avoids VLM finetuning.

% Afterwards, the new prompt augmented with priors, together with visual tokens enhanced by the autoencoder, is then fed into the large language model for reasoning. This combination creates a powerful synergy: \emph{the adapter produces richer visual features that capture the fine-grained characteristics of rare objects, while the text priors guide the language model to accurately interpret these enhanced representations.} Importantly, our work is highly computation-efficient, adapting to different VLM architectures and their limitations, and achieving substantial improvements on rare object reasoning tasks without the need for full model fine-tuning.
\section{Experiments}

\subsection{Experimental Setting}

\noindent \textbf{Datasets.} We conduct experiments on the CODA-LM \cite{chen2025automated} and GeoBench-VLM \cite{danish2025geobench} datasets. For both datasets, we perform recognition and reasoning over referred objects/regions. For the CODA-LM dataset, we train the class embeddings and adapter on the 10,727 QA pairs in the training set and validate on the 1,123 relative QA pairs in the test set. Unlike other autonomous driving VQA datasets, CODA-LM includes novel object classes such as ``stroller'' and ``debris'', as well as novel instances of common objects in the autonomous driving environment, which is rare in large-scale datasets. For the GeoBench-VLM dataset, images are captured by satellite, and the objects/regions are rare classes such as ``storage tank'' and ``roundabout''. Since the classes in the GeoBench-VLM dataset are rare and relative images are hard to collect, we finally train our modules on 361 VQA pairs and validate on a total of 190 samples.

\noindent \textbf{Implementation Details.} We adopt general VLM: LLaVA-1.5-7B/13B as the baseline, which consists of Vicuna-1.5 as the language decoder with a CLIP vision encoder. To demonstrate the generalization of our method, we also test state-of-the-art general vision language models (VLMs) such as Qwen2.5-VL \cite{bai2025qwen2} and InternVL3 \cite{zhu2025internvl3}. To train class embeddings, we employ different VFMs like DINOv3 \cite{simeoni2025dinov3}, SAM \cite{kirillov2023segment} as vision encoder (Note that we report the performance with DINOv3 in the main paper except the ablation study in the Appendix) and CLIP \cite{radford2021learning} as text encoder for feature extraction. For visual adapter, the cross-attention module includes a multi-head transformer with 8 heads. The transformer module contains a single attention layer with an embedding dimension of 1024. Note that we only refine the vision tokens $\mathbf{V}$ in the \textbf{first} decode layer (e.g., total 32 layers for LLaVA) in VLM and report performance in the main paper. The ablation study for refinement in other layers is shown in the Appendix. For model training, all proposed modules are trained using AdamW \cite{loshchilov2017decoupled} optimizer with a learning rate of 1e-4, and a weight decay of 0.01. We train the class embeddings for 20 epochs with a batch size of 128, and finetune the adapter for the VQA task for 10 epochs with a batch size of 1. For the hyperparameter, $\kappa$ is set as 0.95, and $k$ is set as 3 for both datasets. All experiments are conducted on a single RTX 4090 GPU.

\begin{table*}[t]
\centering
\caption{
GPT score comparison on the GeoBench-VLM dataset. 
``+ Ours'' denotes our parameter-efficient refinement built on frozen baseline VLMs. 
The model marked with $^\dagger$ is finetuned on GeoBench-VLM, and models marked with $^\ddagger$ are training-free methods. 
}\vspace{-2mm}
\begin{tabular}{c|cccccc}
\Xhline{1pt}
\textbf{Model / Metrics}
& \textbf{Aerial}$\uparrow$
& \textbf{Maritime}$\uparrow$
& \textbf{Vehicle}$\uparrow$
& \textbf{Sports}$\uparrow$
& \textbf{Construction}$\uparrow$
& \textbf{All}$\uparrow$ \\
\Xhline{1pt}

LLaVA-1.5-7B \cite{liu2023visual}             & 16.5 & 29.8 & 14.0 & 15.5 & 12.2 & 20.9 \\
\rowcolor{blue!8}
LLaVA-1.5-7B + Ours                           & 21.5 & 49.4 & 12.0 & 34.5 & 10.0 & 33.2 \\
Qwen2.5-VL-7B \cite{bai2025qwen2}            & 18.0 & 33.5 & 26.5 & 27.9 & 13.3 & 27.4 \\
\rowcolor{blue!8}
Qwen2.5-VL-7B + Ours                         & \underline{26.7} & \underline{52.3} & 34.0 & \underline{36.7} & 14.2 & \underline{39.3} \\
InternVL3-8B \cite{zhu2025internvl3}         & 26.0 & 46.1 & \underline{38.0} & 35.2 & \underline{20.0} & 37.4 \\
\rowcolor{blue!8}
InternVL3-8B + Ours                          & \textbf{35.4} & \textbf{61.4} & \textbf{42.1} & \textbf{45.7} & \textbf{24.3} & \textbf{48.2} \\
LLaVA-1.5-13B \cite{liu2023visual}           & 13.5 & 37.1 & 19.5 & 13.4 & 13.3 & 23.7 \\
\rowcolor{blue!8}
LLaVA-1.5-13B + Ours                         & 23.6 & 51.2 & 17.4 & 32.5 & 14.1 & 34.9 \\ \hline

\rowcolor{gray!10}
LLaVA-1.5-7B$^\dagger$ \cite{liu2023visual}
                          & 22.4 & 46.7 & 27.3 & 32.1 & 18.3 & 34.7 \\ \hline

\rowcolor{green!5}
Jiang et al.$^\ddagger$ \cite{jiang2025devils}  
                          & 17.4 & 28.1 & 15.8 & 18.9 & 12.8 & 21.4 \\
\rowcolor{green!5}
ControlMLLM++$^\ddagger$~\cite{wu2024controlmllm} 
                          & 18.1 & 27.8 & 16.7 & 22.8 & 14.3 & 22.5 \\
\Xhline{1pt}
\end{tabular}\vspace{-3mm}
\label{tab:geobench}
\end{table*}

\noindent \textbf{Evaluation Metrics.} For CODA-LM \cite{chen2025automated}, we follow the official evaluation protocol to assess performance across different categories. Among these categories, ``Barrier'', ``Other'' and ``VRU'' appear less frequently than the others in CODA-LM. Model performance is measured using the GPT score, which quantifies the semantic similarity between generated answers and ground-truth annotations on a scale of 1 to 100. For GeoBench-VLM \cite{danish2025geobench}, we summarize the object classes into different categories-``Aerial'', ``Maritime'', ``Vehicle'', ``Sports'', and ``Construction'', with ``Sports'' and ``Construction'' being the least frequent. The evaluation metrics are similar to the CODA-LM dataset. The class details of the CODA-LM and GeoBench-VLM datasets are shown in the Appendix.  

% \ZD{We need add a summary of comparisons}

\subsection{Comparison Results}

\noindent\textbf{Results on CODA-LM.}
Table~\ref{tab:coda-lm} reports the performance of different methods on the CODA-LM dataset. 
Across all four base VLMs, our method consistently improves the frozen baselines. 
On the overall \textbf{All} metric, our method brings gains of $+26.3$ (from $46.5$ to $72.8$) on LLaVA-1.5-7B, $+7.5$ (from $67.9$ to $75.4$) on Qwen2.5-VL-7B, $+8.8$ (from $65.4$ to $74.2$) on InternVL3-8B, and $+20.4$ (from $50.8$ to $71.2$) on LLaVA-1.5-13B. 
These gains are particularly pronounced in rare and safety-critical categories. 
For example, on top of the LLaVA-1.5-7B baseline, our method improves ``Barrier'', ``Other'' and ``VRU'' by $+29.0$, $+28.1$, and $+15.6$ points, respectively. 
The Qwen2.5-VL-7B variant achieves the best overall performance among frozen baseline models (75.4 on \textbf{All}), and attains the highest scores on several key categories, including ``Barrier'' (79.8), ``Other'' (73.8), and ``Cone'' (91.7).

We further compare with specific finetuned models like CODA-LM \cite{chen2025automated}, MiniDrive \cite{zhang2024minidrive}, and MPDrive \cite{zhang2025mpdrive} in Table~\ref{tab:coda-lm}. 
CODA-LM achieves the highest overall score of 77.7 on \textbf{All}, while MPDrive reaches 76.1. 
Our variant of Qwen2.5-VL-7B attains 75.4, narrowing the gap to CODA-LM to only 2.3 points, and surpasses all finetuned models on several categories, such as ``Barrier'' (79.8 vs.\ 78.7) and ``Cone'' (91.7 vs.\ 86.2). 
This indicates that a lightweight, class-guided adapter on top of frozen VLMs can recover most of the benefits of heavy task-specific finetuning. 
By contrast, training-free methods (Jiang et al. \cite{jiang2025devils} and ControlMLLM++ \cite{wu2024controlmllm}) only bring marginal improvements over the LLaVA-1.5-7B baseline on the \textbf{All} metric (from 46.5 to 46.7/47.0), and are clearly worse than our variants and finetuned models across different categories.

\vspace{1mm}\noindent\textbf{Results on GeoBench-VLM.}
Table~\ref{tab:geobench} summarizes the results on the GeoBench-VLM dataset. 
Our method again consistently improves over all frozen baselines. On the overall \textbf{All} metric, we observe gains of $+12.3$ (from $20.9$ to $33.2$) on LLaVA-1.5-7B, $+11.9$ (from $27.4$ to $39.3$) on Qwen2.5-VL-7B, $+10.8$ (from $37.4$ to $48.2$) on InternVL3-8B, and $+11.2$ (from $23.7$ to $34.9$) on LLaVA-1.5-13B. 
The InternVL3-8B variant achieves the best overall performance with 48.2 on \textbf{All}, and consistently leads in all five categories. For rare object categories ``Sports'' and ``Construction'', our method improves the relative performance of most tested VLMs: $+8.8$/$+0.9$ for Qwen2.5-VL-7B, $+10.5$/$+4.3$ for InternVL3-8B, and $+19.1$/$+0.8$ for LLaVA-1.5-13B. These results confirm that the proposed strategy generalizes well across different architectures and from driving scenes to satellite imagery.

\begin{figure}[t]
  \centering
  \includegraphics[width=\linewidth]{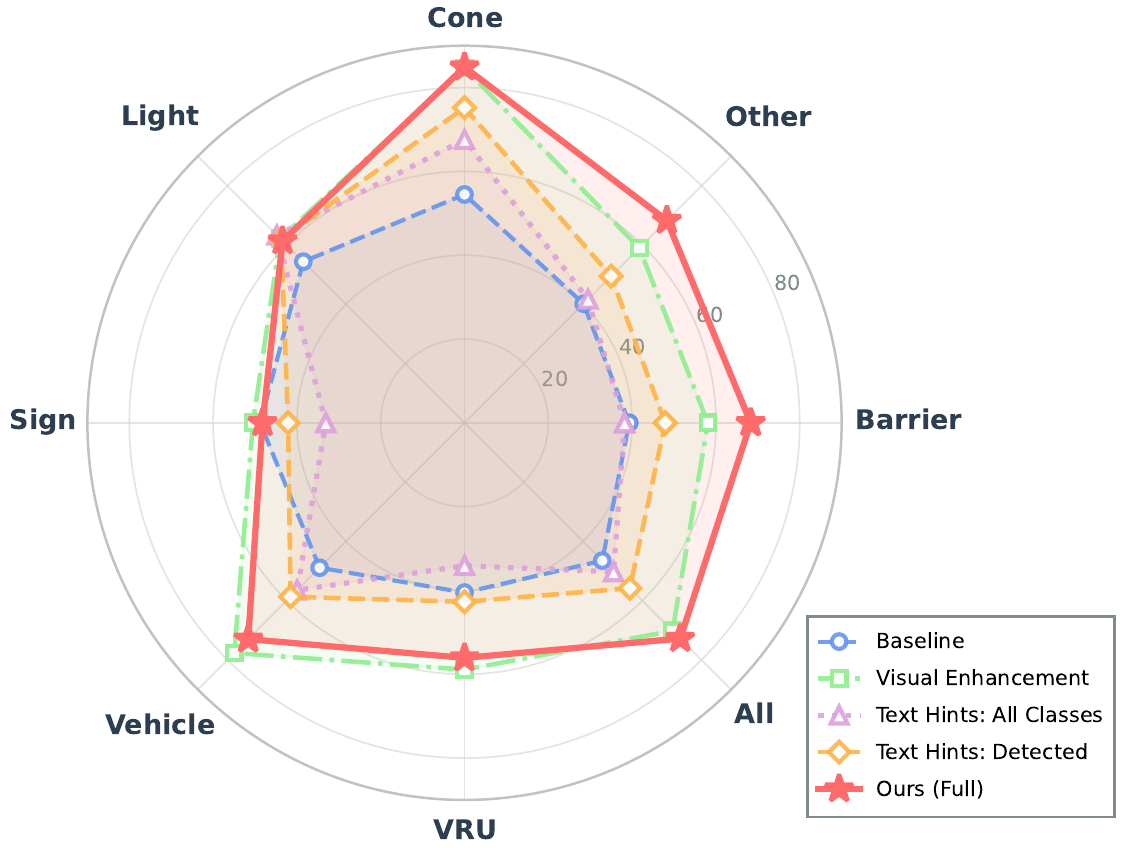}
  \vspace{-6mm}\caption{Ablation study of visual refinement and text hints for LLaVA-1.5-7B on the CODA-LM dataset.}\vspace{-7mm}
  \label{fig:ablation}
\end{figure}

On GeoBench-VLM, a LoRA-finetuned LLaVA-1.5-7B model reaches 34.7 on \textbf{All}, slightly better than our LLaVA-1.5-7B variant (33.2), but clearly worse than our Qwen2.5-VL-7B and InternVL3-8B variants (39.3 and 48.2, respectively). Such marginal improvement is mainly caused by the extremely scarce data in the GeoBench-VLM dataset, which doesn't well finetune the LLaVA. Training-free methods again provide only modest improvements over the LLaVA-1.5-7B baseline (e.g., 21.4 and 22.5 vs. 20.9 on \textbf{All}) and remain far behind our best models.

\subsection{Ablation Study}

% \begin{table}[t]
% \centering
% \caption{Ablation study of visual refinement and text priors for LLaVA-1.5-7B on the CODA-LM dataset. \ZD{Too small font size. Can we use figures instead?}}
% \scalebox{0.6}{
% \begin{tabular}{c c|cccccccc}
% \Xhline{1pt}
% \multirow{2}{*}{\textbf{Visual}} & \multirow{2}{*}{\textbf{Text Priors}} 
% & \multicolumn{8}{c}{\textbf{Metrics}} \\ \cline{3-10}
% & & \textbf{Barrier} & \textbf{Other} & \textbf{Cone} & \textbf{Light} & \textbf{Sign} & \textbf{Vehicle} & \textbf{VRU} & \textbf{All} \\ 
% \Xhline{1pt}
%           & None                & 39.3 & 40.2 & 54.5 & 54.4 & \underline{48.8} & 48.9 & 40.5 & 46.5 \\
% \checkmark& None                & \underline{58.1} & \underline{59.0} & \underline{84.7} & \underline{62.2} & \textbf{50.5} & \textbf{77.7} & \textbf{58.9} & \underline{70.2} \\
%           & All classes         & 38.2 & 41.7 & 67.6 & \textbf{63.3} & 33.1 & 56.5 & 34.1 & 50.3 \\
%           & Detected     & 47.9 & 49.5 & 75.2 & 61.9 & 42.1 & 58.7 & 42.7 & 55.8 \\
% \checkmark& Detected     & \textbf{68.3} & \textbf{68.3} & \textbf{84.9} & 61.4 & 48.2 & \underline{73.0} & \underline{56.1} & \textbf{72.8} \\
% \Xhline{1pt}
% \end{tabular}}
% \label{abaltion-coda}
% \end{table}

\begin{figure}[t]
  \centering
\hspace{-2mm}\includegraphics[width=\linewidth]{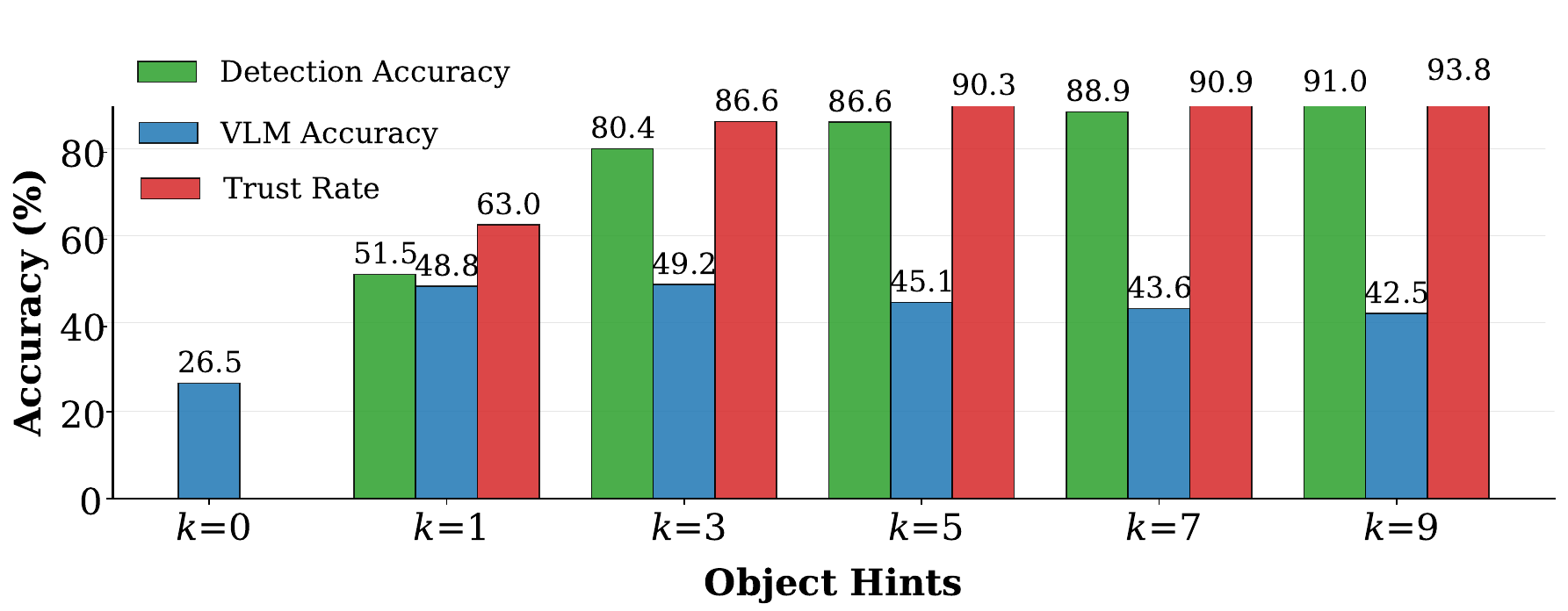}
  \vspace{-2mm}\caption{Comparison of different $k$ for LLaVA-7B on CODA-LM. ``Detection Accuracy'' is the top-$k$ detection accuracy of multi-modal class embeddings for objects. ``VLM Accuracy'' measures how VLMs recognize objects with/without our hints. ``Trust Rate'' is the ratio of VLMs' output that aligns with our hints.}\vspace{-6mm}
  \label{fig:topk_priors}
\end{figure}

\noindent\textbf{Effect of Visual Refinement and Text Hints.} Figure \ref{fig:ablation} demonstrates the effectiveness of our visual token refinement and text hints injection on LLaVA-1.5-7B. In detail, the frozen LLaVA baseline achieves 46.5 on \textbf{All} object classes. Enabling only ``Visual Enhancement'' yields a large improvement of 70.2 (+23.7), confirming that the visual adapter can effectively inject class-discriminative cues into visual tokens. We then examine ``Text Hints'' without visual refinement and compare two construction strategies. Simply appending all available class labels—``All Classes'' as object hints yields 50.3 on \textbf{All} (+3.8) but introduces noise that hurts some categories (e.g., \textbf{Sign}: 48.8$\rightarrow$33.1; \textbf{VRU}: 40.5$\rightarrow$34.1). In contrast, using our ``Detected (Top-$k$)'' object candidates achieves 55.8 on \textbf{All} (+9.3) and provides targeted benefits (e.g., \textbf{Barrier}: 39.3$\rightarrow$47.9; \textbf{Other}: 40.2$\rightarrow$49.5). Finally, combining both components—``Ours (Full)'' provides the best result 72.8 on \textbf{All} (+26.3), and consistently expands the similar improvement on rare categories: \textbf{Barrier}: 39.3$\rightarrow$68.3; \textbf{Other}: 40.2$\rightarrow$68.3; \textbf{VRU}: 40.5$\rightarrow$56.1. These trends show clear complementarity: visual refinement enhances object-level strength in the image tokens, while selective text hints steer the decoder toward the correct regions/labels without introducing prompt noise.

\noindent\textbf{Number $k$ of Injected Object Hints.}
Figure~\ref{fig:topk_priors} further analyzes how many detected classes should be injected as hints. Here, we use accuracy rather than GPT score to more clearly demonstrate the detection accuracy of our multi-modal class embeddings (described in Sec 3.5) and the VLM--LLaVA prediction accuracy on referred objects. Here we don't refine visual tokens in LLaVA but with text hints. As $k$ increases from 1 to 9, the detection accuracy and the VLM's trust rate (the ratio of VLM predictions aligned with the injected hints) steadily improve, indicating that our class embeddings yield increasingly reliable candidates and that VLM prefers our hints. However, VLM's prediction accuracy peaks around $1$--$3$ and then gradually decreases when more hints are added. This indicates that, beyond a certain point, additional candidates start to confuse the model rather than help it. We therefore choose $k=3$ as a good trade-off: it retains the peak VLM accuracy while supplying richer object-level information than a single object.
% For instance, in Figure~\ref{fig:motivation}, providing a long candidate list such as ``bollard, cyclist, bus'' may cause the model to drift from the correct ``bollard'' to the nearby ``cyclist'' due to imperfect visual tokens.

\subsection{Interpretable Analysis}

\begin{figure}[t]
  \centering
  \includegraphics[width=\linewidth]{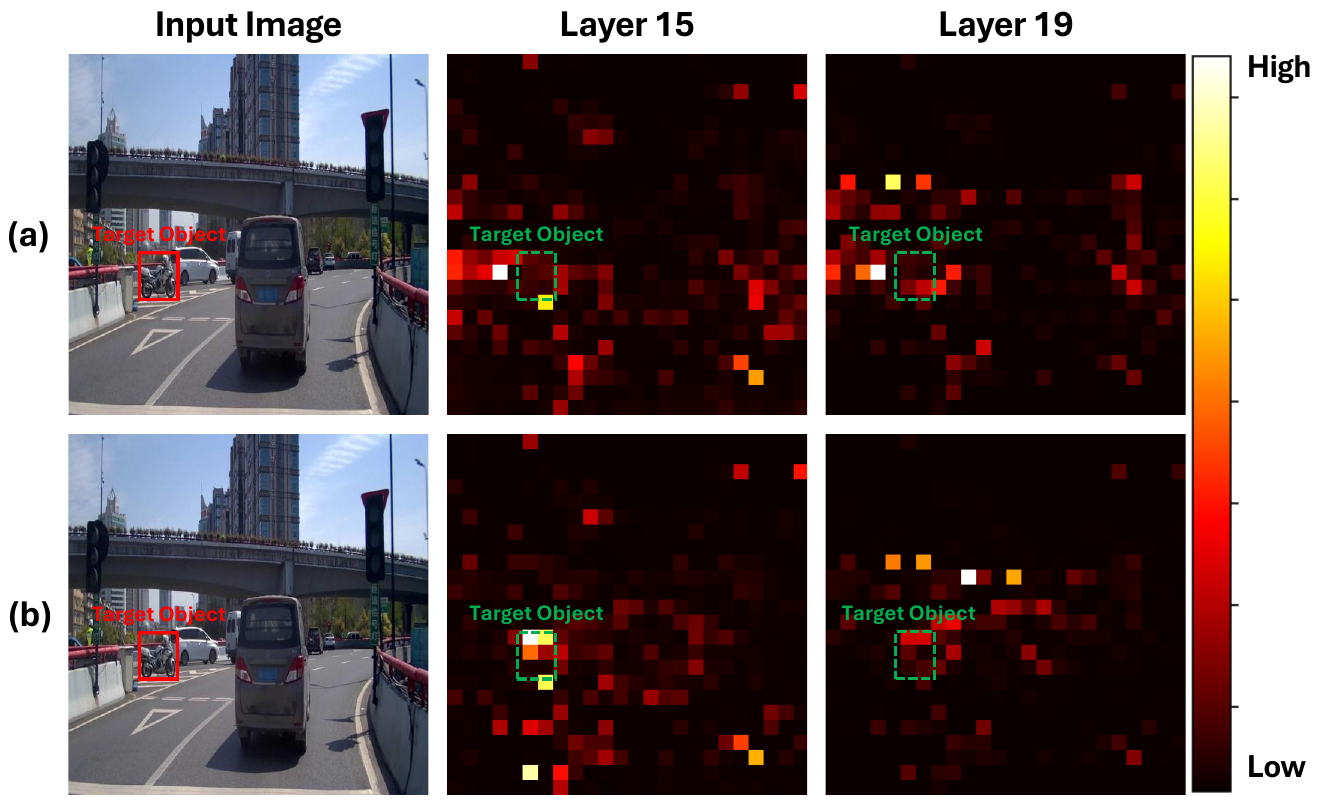}
  \vspace{-7mm}\caption{Attention weights comparison between (a) LLaVA-1.5-7B and (b) LLaVA-1.5-7B + Ours. }
  % Note that bright color means high attention. }
  \label{fig:attention_weight} \vspace{-5mm}
\end{figure} 

\begin{figure}[t]
  \centering
  \includegraphics[width=\linewidth]{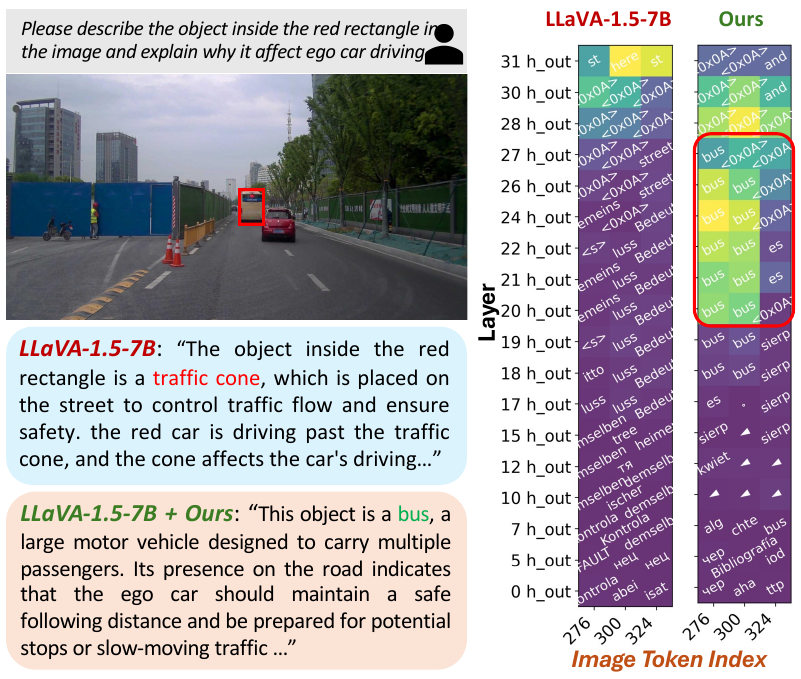}
  \vspace{-7mm}\caption{Interpretation of image hidden states for object class ``bus'' in LLaVA-1.5-7B via logit lens~\cite{jiang2025devils} on CODA-LM dataset. }
  % The heatmap demonstrates that the original LLaVA-1.5-7B shows meaningless retrieved texts in the object regions, and the VLM pays less attention to the objects. Our method improves the semantic information on  image tokens.}
  \label{fig:logit_len} \vspace{-6mm}
\end{figure}

Figure \ref{fig:attention_weight} demonstrates the difference in attention weight from the predicted ``object'' token on the global image tokens. The attention weight quantifies the extent of the ``object'' token's interaction with visual information: a higher attention weight indicates a greater contribution from image tokens during ``object'' token generation \cite{jiang2025devils}. Note that we calculate the average score of multi-head attentions for each layer. To deeply understand the semantic meaning of the refined visual tokens, we adopt the logit lens \cite{jiang2025devils} to probe object hidden states. In VLM, the logit lens takes intermediate hidden states (including visual tokens) and passes them through the same final language head that is normally applied only at the last layer. This decodes each layer's representation into a word distribution over the vocabulary, from which we can see how an image token looks like the object label (e.g., bus, person) and how this semantic prediction evolves across layers. In Figure \ref{fig:logit_len}, the \emph{x}-axis indexes the position tokens corresponding to the referred object ``bus'' in the image, and the \emph{y}-axis is the transformer layers. The heatmap shows which words each object token is assigned by the logit lens across different layers. Compared to the original LLaVA, our refined image tokens are more meaningful for the object class ``bus'', and brighter regions indicate greater confidence in the object label, revealing that our refined tokens provide stronger, more spatially coherent evidence than the baseline.

% In the figure, (a) is the distribution of attention weights from original LLaVA-1.5-7B, and (b) shows the attention from the version after our refinement. The results exhibits that the target object region attracts more attention after our refinement, which mitigates the problem as discussed in Introduction. 

% To deeply understand the semantic meaning of refined image token, we utilize the logit lens proposed by \cite{jiang2025devils} to analyze object hidden states. In VLM, logit lens takes the intermediate hidden states and runs them through the same final language head that is normally used only at the last layer. This method decodes each layer’s representation into a word distribution, so it can be known from which layer an image token already \textbf{looks like} a specific label (e.g., bus, person) and how that semantic prediction evolves across layers. In Figure \ref{fig:logit_len}, the x-axis indexes the position tokens corresponding to the referred object ``bus'' in the input image, y-axis shows the 

% As shown in Figure \ref{fig:logit_len}, LLaVA-1.5-7B exhibits meaningless semantics and less attention on the object region from the input image. The results proves that the visual hidden state in LLaVA-1.5-7B is not only weak but also attracts insufficient attentions. 

% \ZD{Please discuss more details}

% \ZD{I moved the logits len here, you can add more attention weight results as Dr. Hamm suggests.}

% \begin{figure}[t]
%   \centering
%   \includegraphics[width=\linewidth]{image/attention_analysis.pdf}
%   \caption{\ZD{We need to adjust the bb.}Attention analysis between baseline and our method. (a) is the attention weight from our method; (b) shows the attention weight from the baseline; (c) demonstrates the difference of attention weights. Red color means more attention on the visual tokens. \JH{Object class?}}
%   \label{fig:attention_analysis}
% \end{figure}
% We report the attention weights from different layers on visual tokens between baseline and our method in Figure \ref{fig:attention_analysis}. Note that the attention weight is based on ``object'' $\rightarrow$ ``image'', exhibiting how the generated ``object'' token pays attention to the visual features. The results show that VLMs demonstrates obvious interest on the referred object region with text priors and such phenomenon is more salient in the middle of VLM decoder layers. It proves that more attentions from the object region can improve the VLM performance, which is similar to the observations from recent work \cite{}.
% \JH{Very interesting. Perhaps add more examples in the appendix?}

\subsection{Training Efficiency}

We estimate that fully training our LLaVA-1.5-7B adapter on CODA-LM requires roughly $7.7\times10^{5}$ TFLOPs, of which more than $99\%$ comes from the frozen VLM's forward pass. The computation attributable to our adapter is only about $5\times10^{3}$ TFLOPs  (approximately $0.6\%$ of the total), corresponding to the forward and backward passes of its 33.6M parameters. In terms of memory, the end-to-end training pipeline uses about 16.5\,GB of GPU memory, with our method accounting for 3.5\,GB. In contrast, CODA-LM~\cite{chen2025automated} and MPDrive~\cite{zhang2025mpdrive} backpropagate through LoRA modules attached to the full vision and language stacks, so their gradient computation scales with the entire backbone rather than a lightweight adapter.

% We estimate that a full training on CODA-LM of our adapter for LLaVA-1.5-7B requires about $7.7\times10^{5}$ TFLOPs in total, where more than $99\%$ of the cost comes from the frozen VLM forward pass. The actual training computation associated with our adapter is only about $5\times10^{3}$ TFLOPs (approximately $0.6\%$ of the total), corresponding to the forward and backward passes of the 33.6M-parameters. In terms of memory, running our training pipeline requires about 16.5\,GB of GPU memory, with 3.5\,GB from our method. In contrast, CODA-LM \cite{chen2025automated} and MPDrive \cite{zhang2025mpdrive} backpropagate through LoRA modules attached to the full vision and language modules, so their gradient compute scales with the entire backbone rather than a small adapter.

\noindent\emph{More ablation study about class embeddings and loss functions, and the details and qualitative analysis of CODA-LM and GeoBench-VLM are in the Appendix.}

% \subsection{Qualitative Results}

% \ZD{What is this?}

% \begin{figure*}[t]
%   \centering
%   \includegraphics[width=\linewidth]{image/logit_len.pdf}
%   \caption{}
%   \label{fig:attention_analysis}
% \end{figure*}

\section{Conclusion}

% In this work, we investigate the reasons why VLMs fail on rare, object-centric scenes, and find weak visual tokens and low attention on the relevant regions. To mitigate such limitation, we propose an efficient plug-and-play module with learnable multi-modal class embeddings that work in two ways: (i) visual token refinement via cross-attention and (ii) prompt enrichment with object hints. Keeping the VLMs frozen, our method consistently improves rare-object recognition and reasoning while also enhancing performance on common categories. Future work includes scaling to open-vocabulary settings and reducing inference overhead.

In this work, we investigate why VLMs fail on rare, object-centric scenes and identify two key factors: weak visual tokens and insufficient attention to the relevant regions. To address these limitations, we propose an efficient plug-and-play module with learnable multi-modal class embeddings that operates in two ways: (i) visual token refinement via cross-attention, and (ii) prompt enrichment through object hints. Keeping the VLMs frozen, our method consistently improves rare-object recognition and reasoning, while also enhancing performance on common categories. Future work includes scaling to open-vocabulary settings and reducing inference overhead.

% \ZD{Please also list what we plan to put SM.

% 1. text prompts with various orders of object hints;

% 2. text prompts with top-k objects associated with confidences, may need normalization based on similarity.

% }

% \JH{One possible limitation - works only on closed-vocabulary tasks?}
% \JH{Can the proposed method negatively affect other tasks? It doesn't, right?}
\clearpage
{
    \small
    \bibliographystyle{ieeenat_fullname}
    \bibliography{main}
}


\end{document}